\documentclass[fleqn,10pt]{wlscirep}
\usepackage{graphicx,lineno,hyperref,subcaption}
\title{Deep Semi Supervised Generative Learning for Automated PD-L1 Tumor Cell Scoring on NSCLC Tissue Needle Biopsies}

\author[1]{Ansh Kapil}
\author[1]{Armin Meier}
\author[1]{Aleksandra Zuraw}
\author[2]{Keith Steele}
\author[2]{Marlon Rebelatto}
\author[1]{G\"unter Schmidt}
\author[1,*]{Nicolas Brieu}
\affil[1]{Definiens AG, Munich, 80636, Germany}
\affil[2]{MedImmune LLC, Gaithersburg, MD 20878, USA}
\affil[*]{nbrieu@definiens.com}

\begin{abstract}
The level of PD-L1 expression in immunohistochemistry (IHC) assays is a key biomarker for the identification of Non-Small-Cell-Lung-Cancer (NSCLC) patients that may respond to anti PD-1/PD-L1 treatments. The quantification of PD-L1 expression currently includes the visual estimation of a Tumor Cell (TC) score by a pathologist and consists of evaluating the ratio of PD-L1 positive and PD-L1 negative tumor cells. Known challenges like differences in positivity estimation around clinically relevant cut-offs and sub-optimal quality of samples makes visual scoring tedious and subjective, yielding a scoring variability between pathologists. In this work, we propose a novel deep learning solution that enables the first automated and objective scoring of PD-L1 expression in late stage NSCLC needle biopsies. To account for the low amount of tissue available in biopsy images and to restrict the amount of manual annotations necessary for training, we explore the use of semi-supervised approaches against standard fully supervised methods. We consolidate the manual annotations used for training as well the visual TC scores used for quantitative evaluation with multiple pathologists. Concordance measures computed on a set of slides unseen during training provide evidence that our automatic scoring method matches visual scoring on the considered dataset while ensuring repeatability and objectivity.
\end{abstract}

\begin{document}

\flushbottom
\maketitle

\thispagestyle{empty}

\section*{Introduction}
The programmed death receptor 1 (PD-1) checkpoint protein with its ligand - programmed death ligand 1 (PD-L1) plays a major role in the immune escape of the cancerous tumor cells, i.e. in the inhibition of the human immune system responses \cite{zou2016pd, grigg2016pd}. More precisely, the proliferation and activation of T-cells as well as the production of the cytokine signaling proteins are inhibited by the binding of PD-L1 proteins to (i) the PD-1 receptors of activated T-cells and (ii) to the CD80/B7-1 receptors on T-cells and antigen presenting cell. Immunotherapeutic drugs aim at restoring the ability of immune cells to kill tumor cells by blocking this escape pathway. The role of complementary or companion diagnostics assays is, in this context, to help the identification of patients which are likely to benefit from a checkpoint inhibitor therapy, i.e. patients with high tumor levels of PD-L1 \cite{udall2018pd}. Tumor level of PD-L1 is estimated by a trained pathologist on biopsies, often obtained with small needles, stained with a PD-L1 antibody. For non small cell lung carcinoma (NSCLC), which accounts for $85\%$ of all lung cancer related deaths \cite{cruz2011lung}, the Tumor Cell (TC) score is defined as the frequency of PD-L1 expressing tumor cells, i.e. of tumor cells with specifically stained membrane. The tumor level of PD-L1 is then compared to a specific cut-off value, which sets the negative or positive PD-L1 status of the patient \cite{ratcliffe2017agreement,kim2017pd}. 

There are known challenges to an accurate estimation of the Tumor Cell (TC) score \cite{Ventana2017}. First, PD-L1 staining is not restricted to the membrane of tumor cells: tumor cells with strong cytoplasmic but no membrane staining, immune cells (e.g. macrophage and lymphocyte) as well as necrotic and stromal regions are not included in the score calculation despite possible PD-L1 staining. A challenge specific to visual scoring is moreover the difficulty for any human observer to estimate heterogeneous distribution of cell populations, with positive and negative tumor regions being often spatially inter-mixed. These challenges make TC scoring a subject to some variability among pathologists \cite{tsao2018pd}. In this work, we propose an automatic scoring solution based on image analysis which achieves an accuracy similar to visual scoring while ensuring objective and reproducible scores and could potentially be used as a computer aided system to help pathologists make a better decision. The complexity of the scene and the difficulty of the task naturally lead us to opt for a deep learning-based solution. 

Previous works have shown the ability of deep learning-based methods to solve complex tasks in general image analysis and understanding field \cite{krizhevsky2012imagenet, simonyan2014very, he2015delving, szegedy2015going, he2016deep, szegedy2016rethinking} as well as in the more specific field of digital pathology image analysis \cite{Litjens2016, CruzRoa217, Han2017, Komura2018}. As a first example, Litjens et al. showed in their pioneering study \cite{Litjens2016} the potential of deep learning for the detection of prostate cancer regions and of breast cancer metastasis in lymph nodes on digital images of H\&E stained slides. More specially, two fully supervised convolutional neural networks (CNN) \cite{lecun1998gradient,lecun2010convolutional} were independently trained on the complete manual annotations of 200 and 170 tissue slides respectively. 
Cruz-Roa et al. proposed a similar fully supervised CNN-based approach for the detection of invasive breast cancer region in H\&E stained slides \cite{CruzRoa217}, relying on the annotations of nearly 400 slides from multiple different sites for training. 
Most previous works in the field of digital pathology image analysis build on fully supervised networks, which are trained only on labeled information obtained through very extensive manual annotations. Collecting the necessary amount of annotations is however a well known problem in this field. This is because images with the level of complexity observed in digital pathology can and should only be interpreted by experts with several years of training and experience. This is a key difference to other fields of application of deep learning methods, for instance in robotics, for which the complexity arises more from the number and diversity of the classes than on the ability of untrained humans to recognize these classes. Because pathologists can disagree on the interpretation of a given slide, it is often necessary to collect manual annotations of the same slide from different pathologists. While being often necessary in order to reduce ambiguity in the training set, annotating the same slide multiple times significantly increases the burden of manual annotation.

To bypass the need of manual annotation for region segmentation or object detection, some recent works proposed to directly infer the patient label. Bychkov et al.\cite{Bychkov2018} introduced a long short term memory (LSTM) network to directly predict patient outcome on tissue microarrays. Campanella et al. \cite{campanella2018terabyte} developed a multiple instance learning (MIL) solution to predict prostate cancer diagnosis on needle biopsies, the training of the system requiring a huge dataset of more than 12000 slides. The use of needle biopsies and the small size of datasets available in clinical trial studies  make however the use of the aforementioned weakly supervised learning approaches challenging in our scenario. 

As previously shown by Vandenberghe et al. \cite{Vandenberghe2017} for the Her2 scoring of breast cancer slides, scores do not have to be learned and can be accurately replicated using the heuristic definition given in the scoring guidelines and an intermediate detection step. To keep the intermediate detection step while reducing the amount of manual annotation, we propose a semi-supervised learning solution. These approaches still employ manual annotations but make use of raw unlabeled data to lower the necessary amount of labeled data \cite{zhang2016augmenting}. Only a few previous works \cite{Peikari2018, kieffer2017, mormont2018} have used semi-supervised learning for digital pathology image analysis. Peikari et al. recently introduced a cluster-then-label method based on support vector machine classifier that is shown to outperform classical supervised classifiers\cite{Peikari2018}. Sparks et al. proposed an image query approach based on semi-supervised manifold learning \cite{Sparks2016}. Other works focus on transfer learning where the model weights are initialized on other classification task \cite{kieffer2017, mormont2018} or learned on raw unlabeled data using representation learning, the labeled data being then used for model refinement only. Given the recent advances in the field of generative adversarial networks (GAN), our work build on class auxiliary generative adversarial networks (AC-GANs). To the best of our knowledge, this study introduces the first application of AC-GAN networks for digital pathology image analysis as well as the first computer-aided diagnostic tool for PD-L1 scoring on needle biopsies. 

\section*{Methods}
\begin{figure}[t!]
\centering
\includegraphics[width=0.99\textwidth]{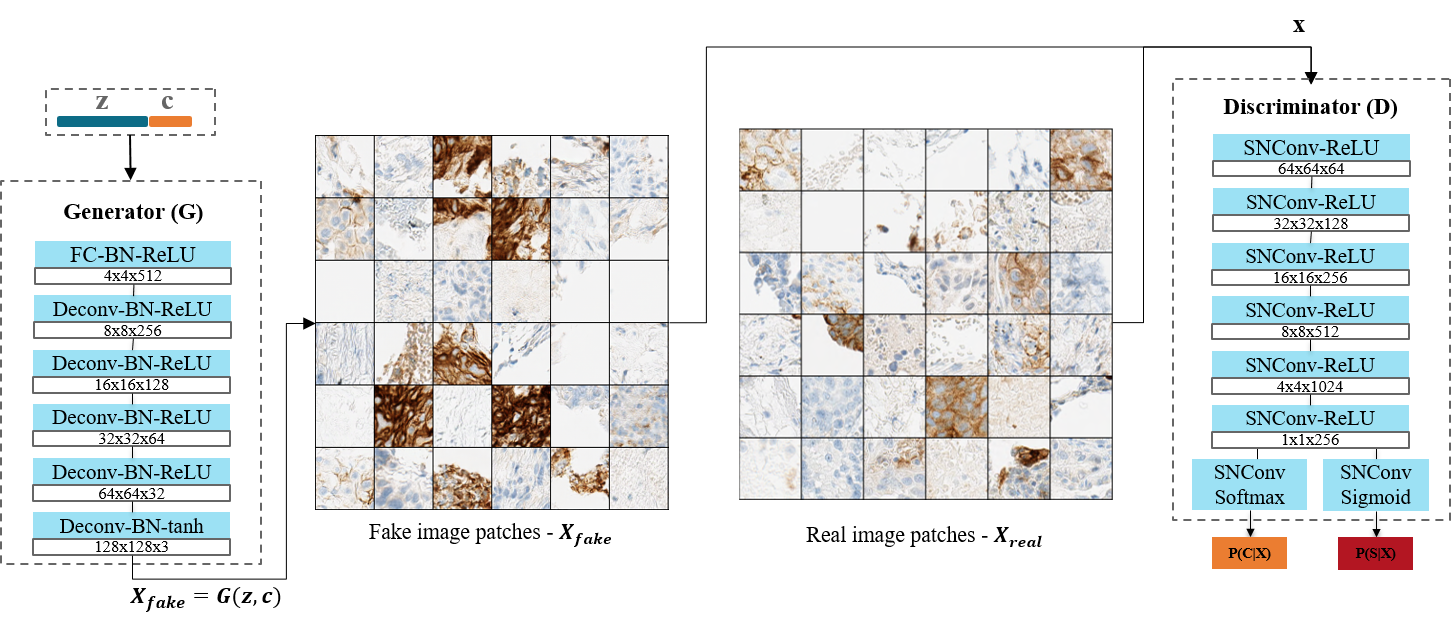}
\caption{Proposed AC-GAN architecture. The noise $z$ and the class one-hot encoding $c$ are concatenated and sent to the generator (G). The generator creates class-conditioned fake images. The classes $c$ and the source information $s$ are jointly predicted by the discriminator (D) given a fake (left) or a real (right) input. $SNConv$ implies convolutional layers using spectral normalization \cite{miyato2018spectral}, which introduces a minor modification to the original AC-GAN architecture \cite{odena2016conditional}.}
\label{fig:acgan}
\end{figure}

The proposed TC scoring algorithm consists of two main steps. First, positive tumor cell regions TC(+) and negative tumor cell regions TC(-) are detected using a deep semi-supervised architecture trained on both labeled and unlabeled data. An Auxiliary Classifier Generative Adversarial Network (AC-GAN) \cite{goodfellow2014} is more precisely chosen to this end.  Second, the TC score is computed as the ratio between the pixel count (i.e. the area) of the detected positive tumor cell regions to the pixel count of all detected tumor cell regions. Since approaches based on pixel counts often show higher performance than cell-count based quantification \cite{bug2017analyzing} and enable an easier annotation workflow, we estimate the TC scores from the pixel counts of PD-L1 positive and negative tumor regions. 

\paragraph{Dataset consolidation for region detection} A small subset of slides is selected and partially annotated by two pathologists for positive tumor cells TC(+), negative tumor cells TC(-), positive lymphocytes, negative lymphocytes, macrophages, necrosis and stromal regions. A simple detection of the tissue and background regions based on Otsu thresholding \cite{Otsu1979} and morphological filtering is performed, leading to the introduction of another class corresponding to non-tissue regions. Labeled image patches are generated on a regular grid defined on the annotated regions that are concordant between the two pathologists. This leads regions with different classes to be discarded from the analysis. On the remaining set of non-annotated slides, unlabeled patches are generated on a regular grid defined on the detected tissue. 

\paragraph{Auxiliary Classifier Generative Adversarial Networks (AC-GAN)}
Let's first recall the principle and architecture of Generative Adversarial Networks introduced by Goodfellow et al. \cite{goodfellow2014}. GANs consist of two neural networks, a generator network (G) and a discriminator network (D). The network G transforms a noise vector $z$, which is sampled from a simple distribution such as an uniform or normal distribution, into an fake image $X_{fake}=G(z)$ using a series of deconvolution and activation layers. The network D classifies an input image as real or fake. More formally, the discriminator outputs the probability distribution over the sources $P(S|X), S\in \lbrace real, fake \rbrace$, through a series of convolutional and activation layers by maximizing the log-likelihood of the correct source:
\begin{equation}
L_{S} = \mathbb{E}\left[log P(S=real|X_{real})\right] + \mathbb{E}\left[log P(S=fake|X_{fake})\right]
\label{eq:1}
\end{equation}
The two networks are trained in opposition following a minimax game \cite{russell2016artificial} formally defined as follows:
\begin{equation}
\min\limits_{G}\max\limits_{D} V(D, G) = \mathbb{E}_{X \sim P_{data}}\left[\log D(X)\right] + \mathbb{E}_{z \sim noise} \left[ \log (1-D(G(z))) \right]
\label{eq:2}
\end{equation}
More intuitively, the discriminator is trained to differentiate whether the image is coming from real image distribution or fake image distribution from G. In the opposition, G is trained to produce images which are more and more difficult to be identified by the discriminator as real or fake images. When the optimum of this minimax game is reached, the generator creates images so close to the real images that they cannot be differentiated by the discriminator. 

The GAN architecture is, however, strictly unsupervised and generative in nature i.e. it can only be used to generate realistic new samples, leading its current main application in digital pathology to be stain normalization \cite{cho2017neural, Shaban2018}. Some recent works \cite{odena2016conditional,chen2016infogan} in the computer vision community proposed its extension to the AC-GAN variant. The AC-GAN leverages the side information on the class labels by two means. First, the generator is conditioned with class labels to  perform conditional image generation: the generator network takes as input a noise vector concatenated with the one-hot embedded class labels. The concatenated vector goes through a series of transformations by a CNN to create fake images $X_{fake}=G(c,z)$. Second, the discriminator, in addition to predicting the correct source probability $P(S|X)$, also performs class label recovery. The discriminator contains an auxiliary classifier network which outputs the probability distribution of the class $P(C|X)$. The network parameters between the auxiliary classifier and the discriminator are shared, which enables joint learning of $P(C|X)$ and $P(S|X)$. This requires the introduction of a second cost function as the log-likelihood of the correct class:
\begin{equation}
L_{C} = \mathbb{E}\left[log P(C=c|X_{real})\right] + \mathbb{E}\left[log P(C=c|X_{fake})\right].
\label{eq:3}
\end{equation}
Similar to the vanilla GAN formulation, the two networks D and G are trained by a minimax game with slight modification. The discriminator D is trained to maximize $L_{C}+L_{S}$ and the generator G to minimize $L_{S}-L_{C}$. Since the maximization and minimization problems are interlaced, the AC-GAN model is trained by alternatively updating the two models G and D. The exact AC-GAN architecture used in this work is displayed in Fig.\ref{fig:acgan}. 

In our application, we are only interested in the classification performance of the network. While we investigate the discriminator performance in more detail, further quantitative assessment of generator has not been performed. Some qualitative examples of images produced by the generator are however displayed in Fig.~\ref{fig:acgan}.

\paragraph{Fully-supervised, non-generative and generative semi-supervised networks}
We test the semi-supervised generative AC-GAN architecture against two baseline classification networks for fully-supervised learning and two baseline non-generative networks for semi-supervised learning. The two chosen fully-supervised architectures, the inception network v2 \cite{SzegedyVISW15} and a shallow VGG \cite{biglenet} network modified to be fully-convolutional and to include batch-normalization \cite{ioffe2015batch} are commonly employed for the analysis of digital pathology images. The two non-generative semi-supervised networks are vanilla auto-encoder networks \cite{hinton2006reducing} built on the two aforementioned classification networks


\section*{Experiments and Results}
\begin{figure}[t!]
\centering
\includegraphics[width=0.99\textwidth]{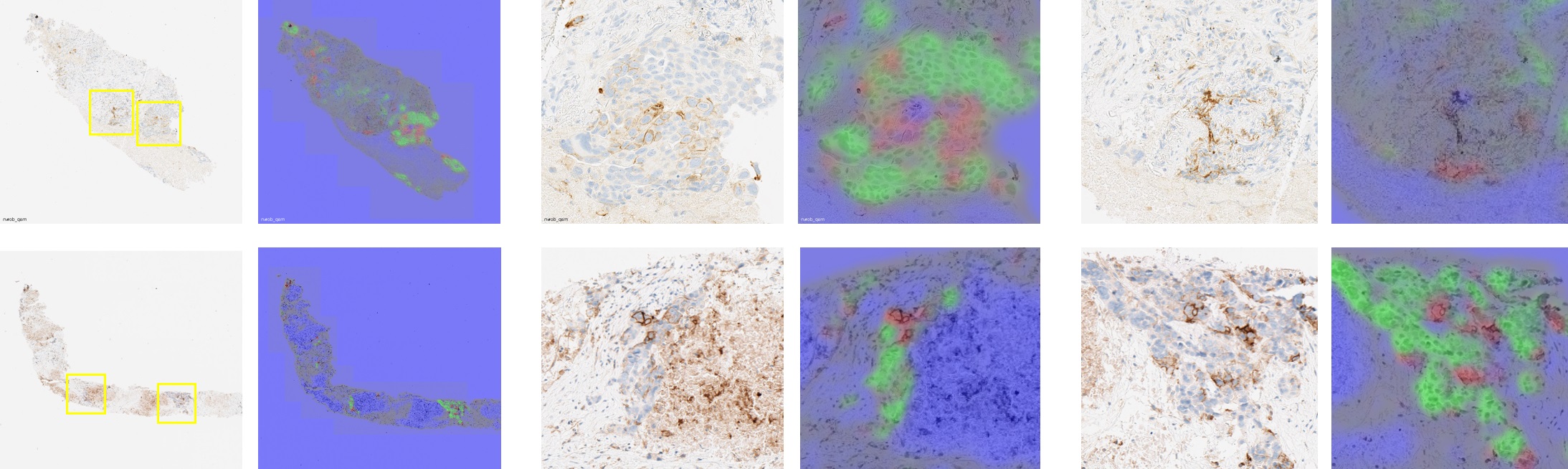}
\caption{Example of class probability maps obtained with AC-GAN model. Left: Original images. Right: Predicted probabilities associated to positive tumor cell region TC(+) (red channel), negative tumor cell region (green channel) and to other classes (blue channel) superimposed to the original grayscale layer. The two yellow boxes overlaid to the far left images correspond to the regions of interest displayed on the right.}
\label{fig:post}
\end{figure}
\begin{figure}[t!]
    \centering
    \includegraphics[width=0.75\textwidth]{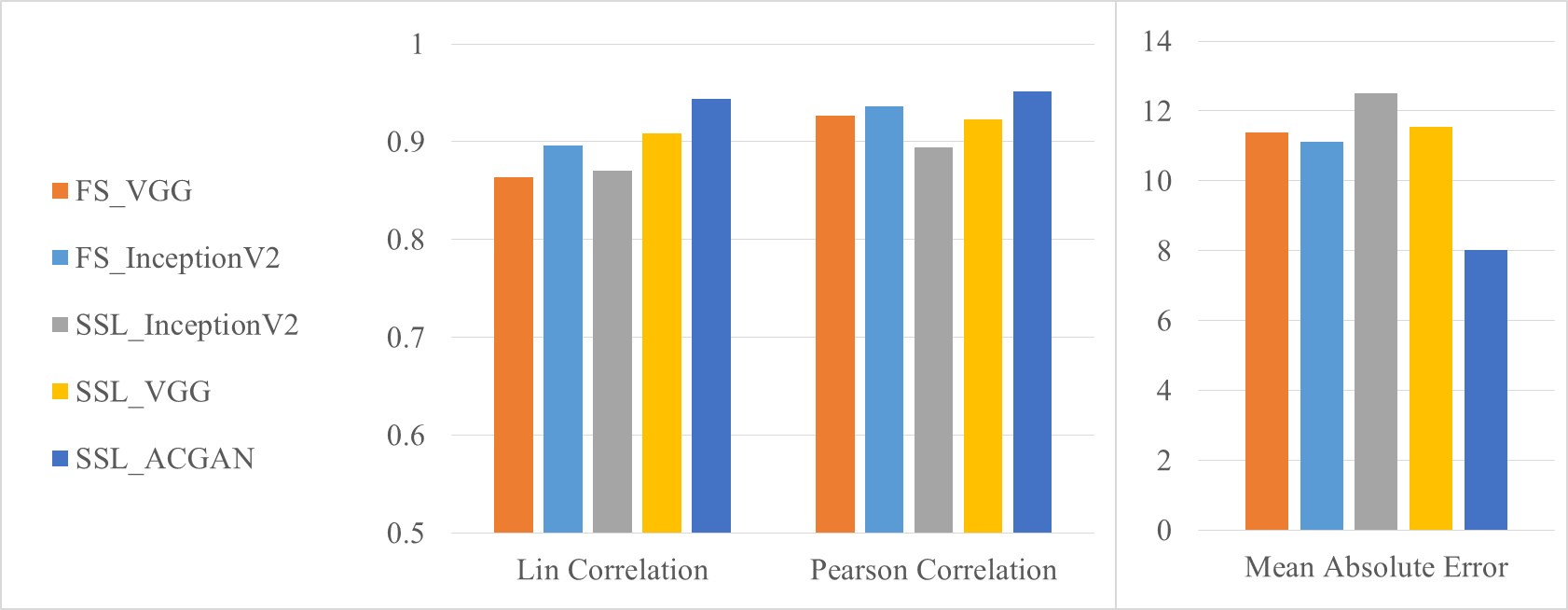}
		\caption{Concordance measures between the consolidated visual TC score and the TC scores automatically estimated from the tumor regions detected using the different supervised and semi-supervised network architectures: fully-supervised shallow VGG net (FS-VGG), fully-supervised inception net v2 (FS-InceptionV2), semi-supervised VGG net (SSL-VGG), semi-supervised inception net v2 (SSL-InceptionV2), and semi-supervised AC-GAN (SLL-ACGAN). Values are computed on the $N_{test}$ cases unseen during training or testing of the networks.}
		\label{fig:barplot}
\end{figure}
\begin{figure}[t!]
    \centering
    \begin{subfigure}[b]{0.32\textwidth}
        \centering
        \includegraphics[width=0.99\textwidth]{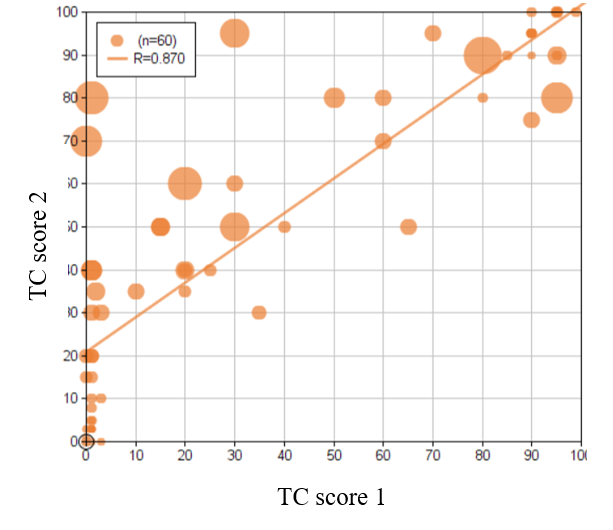}
    \end{subfigure}
		    ~ 
		\begin{subfigure}[b]{0.32\textwidth}
        \centering
        \includegraphics[width=0.99\textwidth]{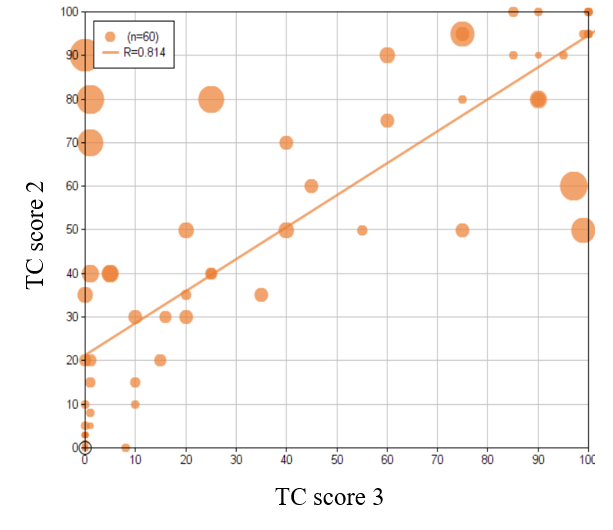}
    \end{subfigure}
		    ~ 
    \begin{subfigure}[b]{0.32\textwidth}
        \centering
        \includegraphics[width=0.99\textwidth]{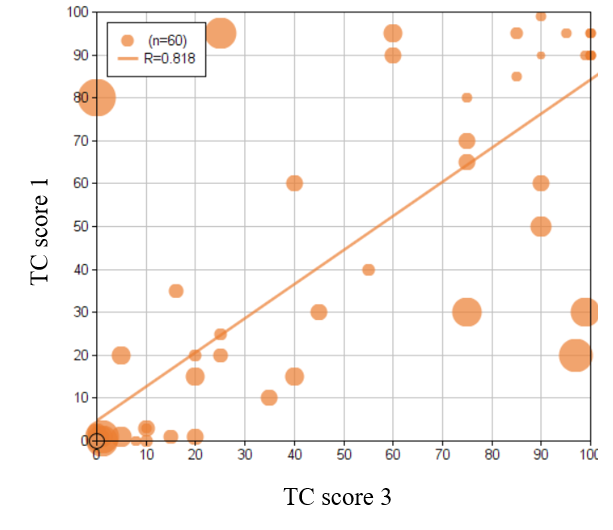}
    \end{subfigure}
		\caption{Pairwise scatter plot between three visual TC scores where each disk represents one of the $60$ visually scored slides. For a given case, the size of the disk is proportional to the inter-rater variability $\Delta_{path}$. }
		\label{fig:InterRater}
\end{figure}
\begin{figure}[t!]
    \centering
    \begin{subfigure}[b]{0.38\textwidth}
        \centering
        \includegraphics[width=0.95\textwidth]{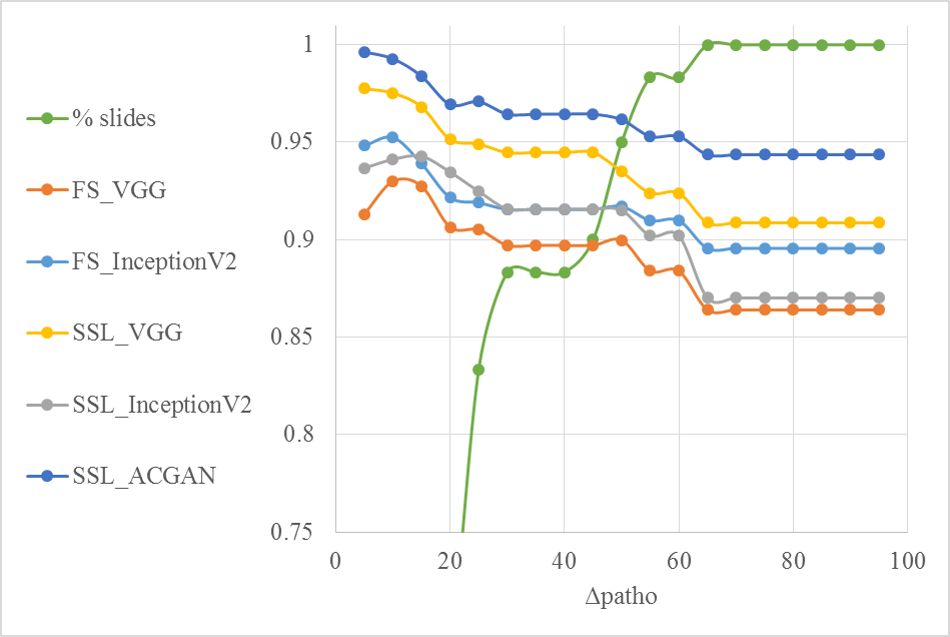}
        \caption{Lin's Correlation Coefficient}
    \end{subfigure}%
    ~ 
    \begin{subfigure}[b]{0.28\textwidth}
        \centering
        \includegraphics[width=0.95\textwidth]{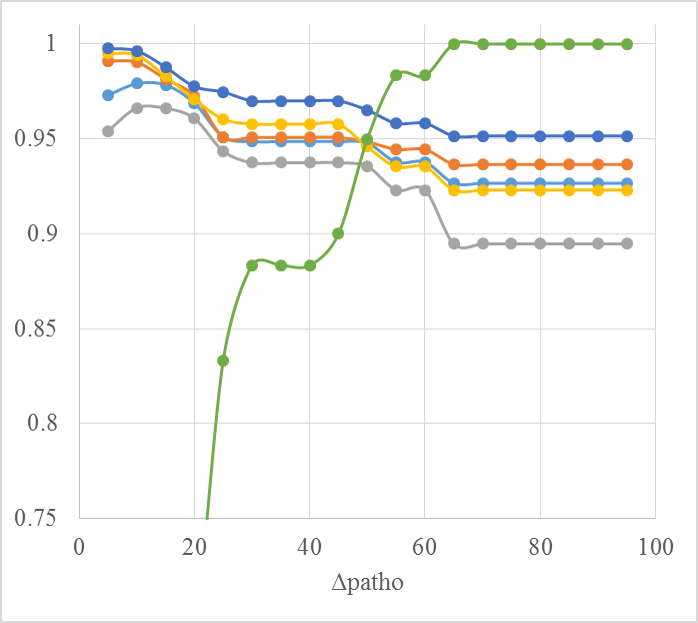}
        \caption{Pearson Correlation Coefficient}
    \end{subfigure}
		~
		\begin{subfigure}[b]{0.28\textwidth}
        \centering
        \includegraphics[width=0.95\textwidth]{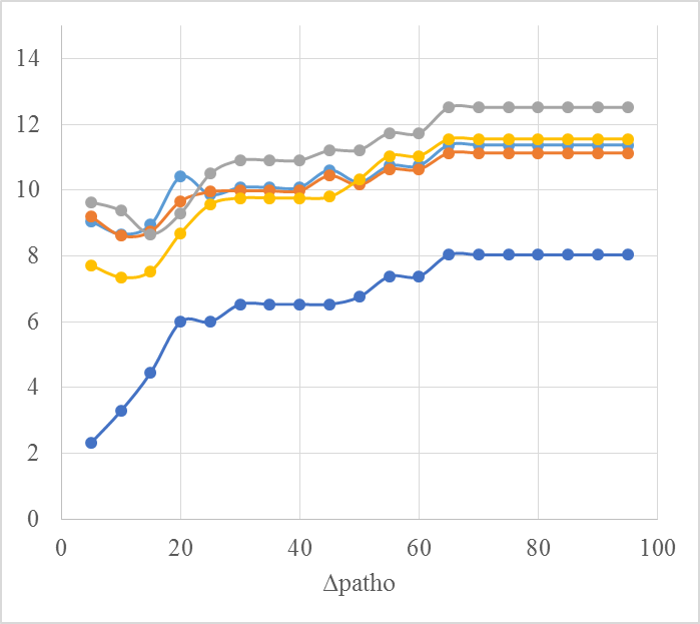}
        \caption{Mean Absolute Error}
    \end{subfigure}
		\caption{Concordance measures between the consolidated visual TC score and the TC scores automatically estimated from the tumor regions detected using the different supervised and semi-supervised network architectures, for increasing maximum levels of inter-rater variability $\Delta_{patho}$ on the x-axis. The green curve indicates the percentage of slides for which the inter-rater variability stays below the given level.}
		\label{fig:LinCurve1}
\end{figure}


\paragraph{Manual annotation and visual TC scoring datasets} Our dataset consists of 270 NSCLC needle biopsy slides from a subset of the clinical trials (NCT01693562) and (NCT02000947). The slides are stained with the Ventana PD-L1 (SP263) assay \cite{rebelatto2016development} and scanned on an Aperio scanner. Scene resolution is $0.198\mu m$ per pixel. On a subset of $N_{score}=60$ slides, visual TC scoring has been performed on the scanned digital slides by two in-house pathologists. This completes the visual TC scoring on glass-slides obtained from an external source (Ventana Medical System Inc.). A smaller subset $N_{cnn}=20$ is selected across the range of TC scores for training and testing the supervised part of the region detection model and have as such been partially manually annotated by the two in-house pathologists using an in-house annotation software. More specifically, $15$ slides are used for training and $5$ slides for testing and optimizing the model parameters. The accuracy of the resulting automated TC score is estimated against the three visual TC scores on the remaining $N_{test}=40$ unseen slides, i.e. unused for training or testing the supervised nor the unsupervised part of the region detection model. 

\paragraph{Model training} 
Patches of size $128\times128$ pixels are extracted on a regular grid of $20$ pixel stride defined on the concordant annotated regions and are augmented using $90$ degree rotation. We sample unlabeled patches on a regular grid of $60$ pixel stride defined on the tissue area of the $210$ slides which have not been scored by all pathologists as well as on the $15$ annotated slides which are used for generating the labeled training database. This yields a total of around 180k labeled and 400k unlabeled patches for training as well as 40k labeled patches for testing. All models are trained using the same patches. Batches with $64$ labeled patches are used for training the fully-supervised networks. Batches with $32$ labeled patches and $32$ unlabeled patches are used for training the three semi-supervised networks. For the two non-generative semi-supervised networks, the reconstruction loss is computed on the complete batch and the classification loss on the labeled patches only. For the AC-GAN, the generative loss is computed on the complete batch and the classification loss on the labeled patches only. Training of the four baseline networks and of the generator and discriminator of the AC-GAN network is performed for $100k$ and $200k$ iterations respectively using the Adam optimizer \cite{kingma2014adam} with the following learning parameters: $lr=0.0001$, $beta1=0.5$, $beta2=0.999$. For each network, we select the model weights that maximize the accuracy on the test dataset in order to avoid overfitting on the training set. The developed framework is based on the open-source Keras API \cite{chollet2015keras} and Tensorflow framework \cite{tensorflow2015-whitepaper}.

\paragraph{Prediction and automated TC scoring} 
The prediction is restricted to the detected tissue regions and performed in a sliding window manner with a stride of $32$ pixels. An example of region detection results is provided in Fig.~\ref{fig:post}. On each of the $N_{score}=60$ slides for which the three visual TC scores are available, we predict the different class probabilities and assign each pixel to the class label of the maximum probability. Given the resulting TC(+) and TC(-) pixels in a given slide, we approximate the corresponding TC score as the ratio of the number of tumor positive cell pixels to the total number of tumor cell pixels:
\begin{equation}
\label{eq:TCscore}
TC_{cnn} = \frac{\#TC(+)}{\#TC(-) + \#TC(+)}.
\end{equation}
We then compute for each slide the consolidated visual score $TC_{ref.}$ as the median of the three visual scores. The following concordance measures between the automated TC score and the consolidated visual score is calculated on the set of $40$ unseen slides only to ensure an independent estimation of the performance: Lin's concordance coefficient (Lcc), Pearson correlation coefficient (Pcc) and Mean Absolute Error (MAE). As reported in Fig.~\ref{fig:barplot}, the AC-GAN achieves on all measures a higher level of agreement to the visual scores ($Lcc=0.94$, $Pcc=0.95$, $MSE=8.03$) than any other fully-supervised and semi-supervised network.

\paragraph{Inter-rater variability of visual TC scoring} 
To quantify the variability between the visual scores (cf. Fig.~\ref{fig:InterRater}), we additionally estimate for each slide the inter-rater variability $\Delta_{path.}$ as the mean absolute pairwise difference between the associated visual scores: 
\begin{equation}
\label{eq:pathDelta}
\Delta_{path.} = \frac{1}{6}\sum_{\substack{1\leq i \leq 3\\ i < j \leq 3}}|TC_{path_i}-TC_{path_j}|.
\end{equation}
Please note that Tsao et al. recently confirmed the very high concordance of PD-L1 scoring on glass slides and digital slides \cite{tsao2018pd}, leading us to pool of all data to estimate inter-rater variability independently of a glass/digital scoring. Given increasing maximum levels of inter-rater variability, we restrict the computation of the aforementioned concordance measures to the slides whose associated value stays below the given maximum. As displayed in Fig.~\ref{fig:LinCurve1}, the automated TC scores become more concordant to the consolidated visual scores the more concordant the visual scores are. A Lin's concordance coefficient of $0.96$ is for example reached on the TC scores obtained with the AC-GAN architecture on cases for which the inter-rater variability is smaller than $40\%$. The better performance of the semi-supervised generative AC-GAN network over the other networks is consistent across highly-concordant and low-concordant cases. 

\paragraph{Automated AC-GAN scoring and visual TC scoring} 
We compare hereby in more detail the performance of the automated score based on the AC-GAN detection to the three visual scores by pathologists. Table~\ref{tab:1} reports the same concordance measures as above between all the visual scores and the AC-GAN score. To study the ability of the proposed automated solution to determine the patient status, we additionally compute  the Overall Percent Agreement (OPA), Negative Percent Agreement (NPA) and Positive Percent Agreement (PPA) at the $25\%$ cut-off (cf. Table~\ref{tab:2}). This cut-off value has been shown to optimize the probability of responses to treatment \cite{rebelatto2016development}.

The TC score computed from the AC-GAN predicted regions yields for all but one case to the highest correlation and the lowest absolute error. Note that in the only case where this does not hold, the Pearson correlation of the automated score $TC_{cnn}=89$ is very close to the highest value $TC_1=0.90$. Computing the concordance to the visual scores $TC_3$ estimated on the microscope, we note that the automated TC scores leads to a higher agreement (Lcc=0.88, Pcc=0.89, MAE=11.3) than the two pathologist scores  estimated on digital slides ($TC_1$:Lcc=0.81, Pcc=0.82, MAE=13.2) - ($TC_2$:Lcc=0.79, Pcc=0.82, MAE=16.6). These observations are confirmed on the reported OPA values of the low/high PD-L1 status at 25\% cut-off. The automated scoring systematically yields the highest agreement with the visual TC score. In particular, considering the third visual TC scoring, automated scoring achieves an $OPA$ of 0.88 to be compared with OPAs of 0.88 and 0.80 observed for the first and second pathologists visually scoring on digital slides .

\begin{figure}[t!]
    \centering
    \begin{subfigure}[b]{0.38\textwidth}
        \centering
        \includegraphics[width=0.95\textwidth]{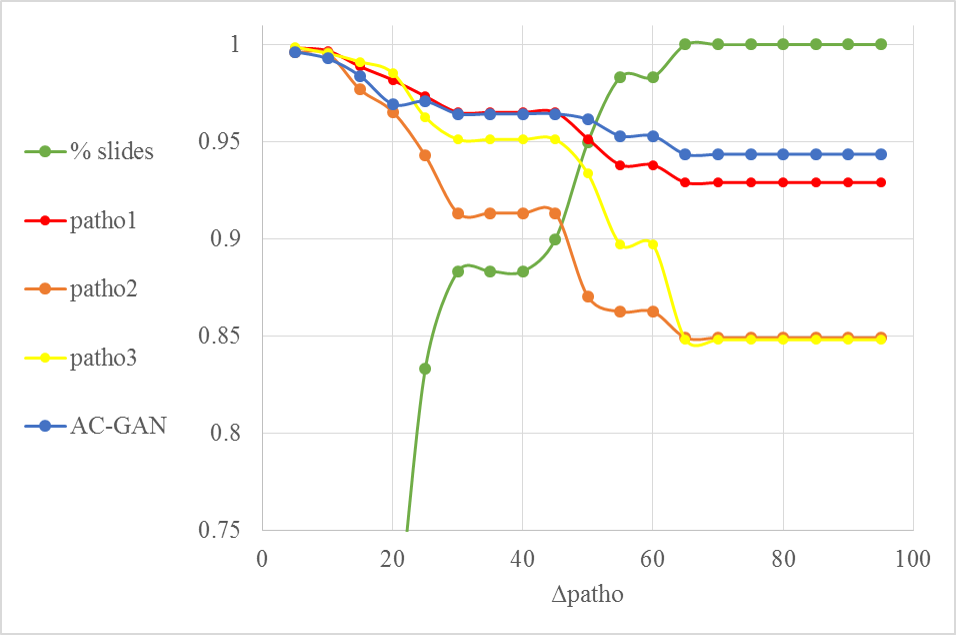}
        \caption{Lin's Correlation Coefficient}
    \end{subfigure}%
    ~ 
    \begin{subfigure}[b]{0.28\textwidth}
        \centering
        \includegraphics[width=0.95\textwidth]{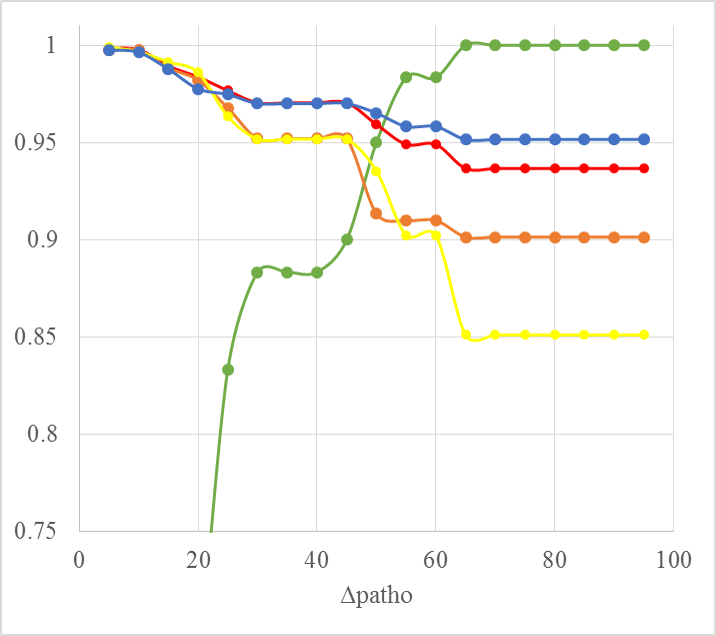}
        \caption{Pearson Correlation Coefficient}
    \end{subfigure}
		~
		\begin{subfigure}[b]{0.28\textwidth}
        \centering
        \includegraphics[width=0.95\textwidth]{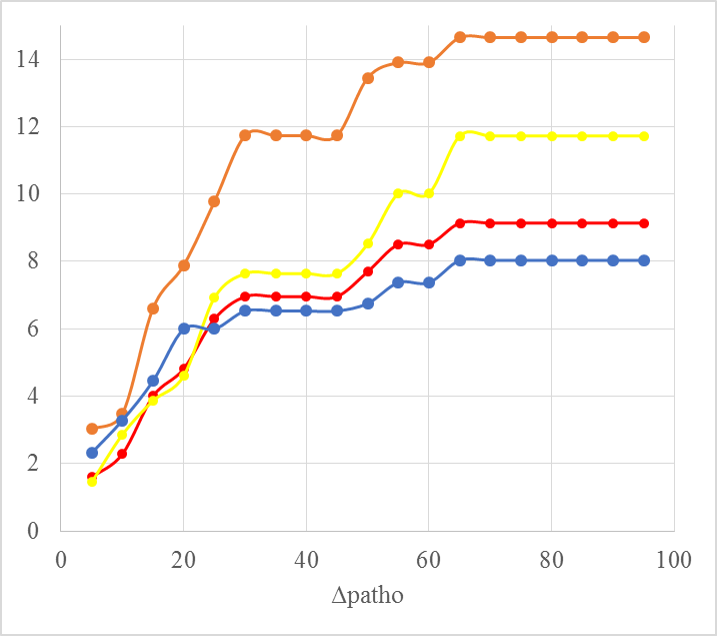}
        \caption{Mean Absolute Error}
    \end{subfigure}
    \caption{Considering the three visual TC scores and the TC score estimated using the region detected by the AC-GAN network, concordance measures between each TC score and the median of the three remaining TC scores for increasing maximum levels of inter-rater variability $\Delta_{path.}$ on the x-axis. The green curve indicates the percentage of slides for which the inter-rater variability stays below the given level.}
		\label{fig:LinCurve2}
\end{figure}

To further analyze the concordance of automated scoring versus visual scoring, we consider the AC-GAN score and the three visual scores independently of their source and compute for each of the four resulting scores all concordance measures between each score and the median of the three remaining scores. Results displayed in Table~\ref{tab:3} and Fig.~\ref{fig:LinCurve2} provide further evidence of the good performance of the automated TC score. In all measures, the automated TC scores systematically outperform the visual scores.

\begin{table}[t!]
\centering
\begin{tabular}{ |c||c|c|c||c|c|c||c|c|c||c|c|c|} 
 \hline
  {}           	& \multicolumn{3}{c}{$TC_1$} 			& \multicolumn{3}{c}{$TC_2$} 			& \multicolumn{3}{c}{$TC_3$} 	& \multicolumn{3}{c}{$TC_{cnn}$}\\
	{}						& Lcc & Pcc & MAE 								& Lcc & Pcc & MAE 								& Lcc & Pcc & MAE 						& Lcc & Pcc & MAE\\ 
 \hline 
 \hline 
 $TC_1$       	& - & - & - 							& 0.84* & 0.90* & 15.3 					& 0.81 & 0.82 & 13.2 						& 0.95* & 0.95* & 7.4*\\ 
 \hline
 $TC_2$        	& 0.84 & 0.90 & 15.3 							& - & - & -  					& 0.79 & 0.82 & 16.6 						& 0.84 & 0.89 & 15.0\\ 
 \hline
 $TC_3$        	& 0.81 & 0.82 & 13.2 							& 0.79 & 0.82 & 16.6 						& -  & - & - 					& 0.88 & 0.89 & 11.3\\ 
 \hline
 $TC_{cnn}$ 		& 0.95* & 0.95* & 7.4* 						& 0.84* & 0.89 & 15.0* 				& 0.88* & 0.89* & 11.3* 				& - & - & -\\
 \hline
\end{tabular}
\caption{Pairwise Lin's concordance / Pearson correlation coefficients and Mean Absolute Error (Lcc/Pcc/MAE) between all visual TC scores and estimated TC score based on AC-GAN. The star indicates for each concordance measure the TC score yielding the best performance, i.e. the maximum value for Lcc and Pcc and the minimum value for the MAE. Values are computed on the $N_{test}$ cases unseen during training or testing of the AC-GAN model.}
\label{tab:1}
\end{table}

\begin{table}[!]
\centering
\begin{tabular}{ |c||c|c|c||c|c|c||c|c|c||c|c|c| } 
 \hline
 {}          	& \multicolumn{3}{c}{$TC_1$} 		& \multicolumn{3}{c}{$TC_2$} 		& \multicolumn{3}{c}{$TC_3$} 		& \multicolumn{3}{c}{$TC_{cnn}$}\\
 {}						& OPA & NPA & PPA 							& OPA & NPA & PPA 							& OPA & NPA & PPA 							& OPA & NPA & PPA\\ 
 \hline 
 \hline 
 $TC_1$       & - &- & -  										& 0.78 & 0.62 & 1.00 						& 0.88* & 0.88 & 0.88* 					& 0.90* & 0.88 & 0.94*\\ 
 \hline		
 $TC_2$       & 0.78 & 1.00* & 0.65						& - & - & -											& 0.80 & 1.00* & 0.69 					& 0.83 & 1.00* & 0.73\\ 
 \hline
 $TC_3$       & 0.88 & 0.91 & 0.83						& 0.80 & 0.65 & 1.00						& - & - & - 										& 0.88 & 0.87 & 0.89\\ 
 \hline
 $TC_{cnn}$ 	& 0.90* & 0.95 & 0.84*					& 0.83* & 0.68* & 1.00*					& 0.88* & 0.91 & 0.84						& - & - & -\\
 \hline
\end{tabular}
\caption{Pairwise Overall, Negative and Positive Percent Agreement (OPA/NPA/PPA) at the 25\% cut-off between all visual TC scores and estimated TC score based on AC-GAN. The star indicates for each concordance measure the TC score yielding the best performance, i.e. the maximum value. Values are computed on the $N_{test}$ cases unseen during training or testing of the AC-GAN model.}
\label{tab:2}
\end{table}

\begin{table}[!]
\centering
\begin{tabular}{ |c||c|c|c|c| } 
 \hline
 -             			& $TC_1$ & $TC_2$ & $TC_3$ & $TC_{cnn}$\\
 \hline 
 \hline 
 Lin / Pearson / MAE & 0.93 / 0.94 / 9.14	&  0.85	/ 0.90 / 14.65 & 0.85	/ 0.85 / 11.73 & 0.94 / 0.95 / 8.00\\
\hline
 OPA / NPA / PPA 		 & 0.85 / 0.95 / 0.76 &  0.78 / 0.63 / 1.0   & 0.85 / 0.90 / 0.80  & 0.88 / 0.90 / 0.85\\
 \hline
\end{tabular}
\caption{Concordance measures between the visual and automated TC scores against the median of the other scores on the $N_{test}$ cases unseen during training or testing of the AC-GAN model.}
\label{tab:3}
\end{table}

\begin{figure}[t!]
    \centering
    \begin{subfigure}[b]{0.38\textwidth}
        \centering
        \includegraphics[width=0.95\textwidth]{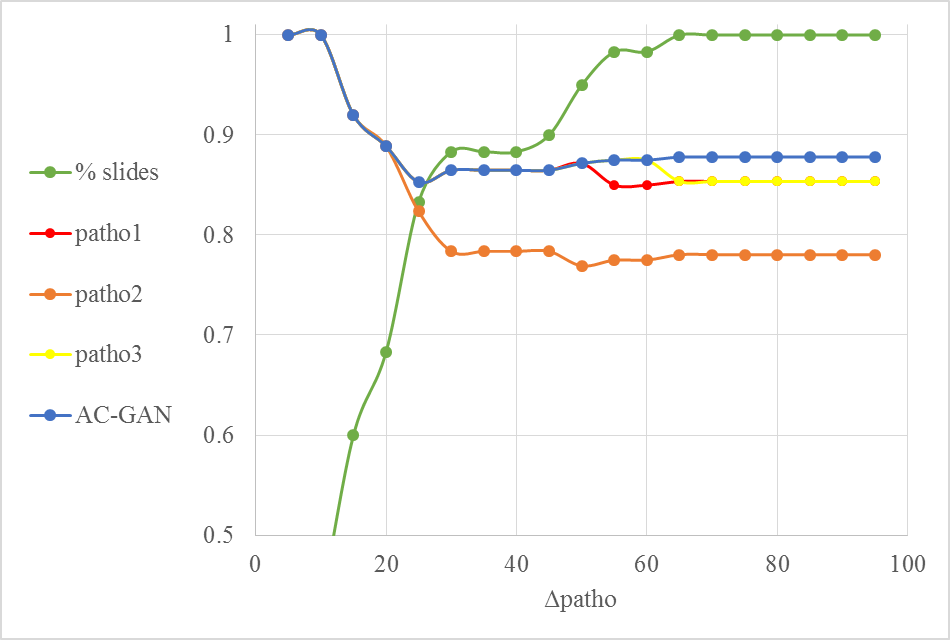}
        \caption{Overall Percent Agreement}
    \end{subfigure}%
    ~ 
    \begin{subfigure}[b]{0.28\textwidth}
        \centering
        \includegraphics[width=0.95\textwidth]{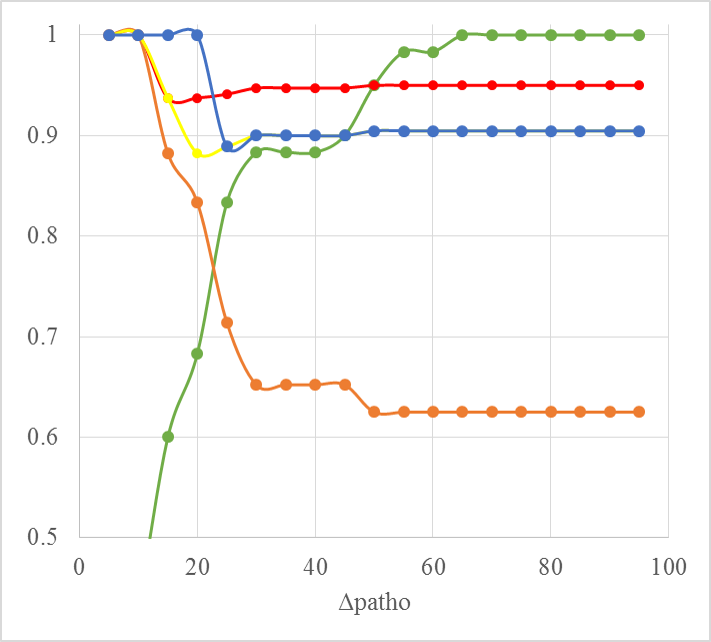}
        \caption{Negative Percent Agreement}
    \end{subfigure}
		~
		\begin{subfigure}[b]{0.28\textwidth}
        \centering
        \includegraphics[width=0.95\textwidth]{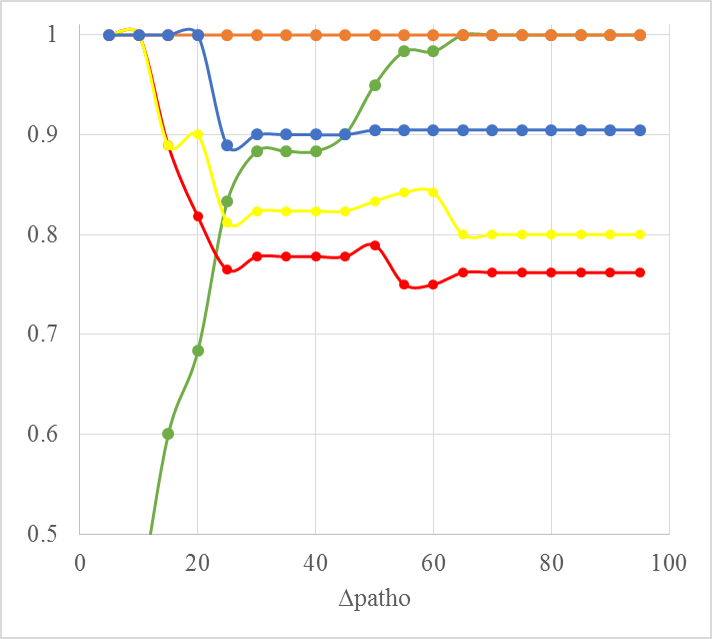}
        \caption{Positive Percent Agreement}
    \end{subfigure}
    \caption{Considering the three visual TC scores and the TC score estimated using the region detected by the AC-GAN network, concordance measures between the patient status estimated from each TC score and patient status estimated from the median of the three remaining TC scores for increasing maximum levels of inter-rater variability $\Delta_{path.}$ on the x-axis. The green curve indicates the percentage of slides for which the inter-rater variability stays below the given level.}
		\label{fig:OPACurve2}
\end{figure}

\section*{Discussion and Conclusion}
The aim of anti-PDL1 therapies is to revive the immune response to cancer cells: inhibiting the PD-L1 pathway reverses T-cell exhaustion and restores T cell's cytotoxic activity. Patients with high expression generally showing higher response rates and longer progression free survival than patients with low expression, an accurate testing of PD-L1 expression may help take the best treatment decision on whether or not to follow such therapy. There is a significant heterogeneity between the existing PD-L1 expression tests, different antibodies being employed as companion and complementary diagnostics for different immunotherapeutic drugs, different patterns of staining (tumor cells only or tumor cells and tumor infiltrating immune cells) being considered for different indications (NSCL resp. urothelial carcinoma) and finally different cutoffs being used for different antibodies for a given indication. 

While the focus of this work is to replicate the test corresponding to the antibody clone SP263 on NSCLC patients, we present what is to our knowledge the first proof of concept study showing that deep learning enables an accurate and automated estimation of the PD-L1 expression level and PD-L1 status on needle biopsies samples. The performed analysis of inter-rater variability suggests that the accuracy achieved by the proposed automated scoring method is concordant with visual scoring by pathologists on our dataset. This work focuses on the automated estimation of the PD-L1 TC score yet, it more generally introduces the first application of deep semi-supervised and generative learning networks (AC-GAN) in the field of digital pathology. It also provides first evidence of the good performance of this model against standard fully supervised learning networks. 

Going beyond the presented proof of concept, we believe that further evidence could be provided by increasing the size of the unseen dataset on which the agreement of the automated and visual TC scores has been computed as well as ensuring that the comparison between the visual scores is not biased by external parameters such as (i) the heterogeneous experience of the pathologists and (ii) if the scoring is performed on digital or glass slides. The presented work also opens the way for PD-L1 scoring in bladder cancer which involves the scoring of both tumor cells and immune cell regions and for which the reliability among pathologists is extremely weak \cite{tsao2018pd}. In general, we envision that the success of checkpoint inhibitor related immune therapies can be increased by automated profiling of tumor and immune cells with respect to their cluster of differentiation (CD) protein expression levels. This study on CD274 (PD-L1) in NSCLC is the first conclusive step in this journey.

\section*{Author contributions statement}
A.K., N.B., A.M., and G.S. designed the analysis. A.K., N.B., A.M. developed the image analysis and statistical analysis components. A.K. and N.B. wrote the manuscript. K.S. and M.R. provided the patient cohort. A.Z. provided the manual annotation and TC scoring of the slides. All authors reviewed the manuscript. 

\section*{Additional information}
A.K., A.M., A.Z., G.S., N.B. are full time or part time employees of Definiens AG. K.S., M.R. are full time or part time employees of Medimmune LLC. 

\bibliography{bibliography}

\begin{thebibliography}{10}
\expandafter\ifx\csname url\endcsname\relax
  \def\url#1{\texttt{#1}}\fi
\expandafter\ifx\csname urlprefix\endcsname\relax\def\urlprefix{URL }\fi
\expandafter\ifx\csname doiprefix\endcsname\relax\def\doiprefix{DOI }\fi
\providecommand{\bibinfo}[2]{#2}
\providecommand{\eprint}[2][]{\url{#2}}

\bibitem{zou2016pd}
\bibinfo{author}{Zou, W.}, \bibinfo{author}{Wolchok, J.~D.} \&
  \bibinfo{author}{Chen, L.}
\newblock \bibinfo{journal}{\bibinfo{title}{Pd-l1 (b7-h1) and pd-1 pathway
  blockade for cancer therapy: Mechanisms, response biomarkers, and
  combinations}}.
\newblock {\emph{\JournalTitle{Science translational medicine}}}
  \textbf{\bibinfo{volume}{8}}, \bibinfo{pages}{328rv4--328rv4}
  (\bibinfo{year}{2016}).

\bibitem{grigg2016pd}
\bibinfo{author}{Grigg, C.} \& \bibinfo{author}{Rizvi, N.~A.}
\newblock \bibinfo{journal}{\bibinfo{title}{Pd-l1 biomarker testing for
  non-small cell lung cancer: truth or fiction?}}
\newblock {\emph{\JournalTitle{Journal for immunotherapy of cancer}}}
  \textbf{\bibinfo{volume}{4}}, \bibinfo{pages}{48} (\bibinfo{year}{2016}).

\bibitem{udall2018pd}
\bibinfo{author}{Udall, M.} \emph{et~al.}
\newblock \bibinfo{journal}{\bibinfo{title}{Pd-l1 diagnostic tests: a
  systematic literature review of scoring algorithms and test-validation
  metrics}}.
\newblock {\emph{\JournalTitle{Diagnostic pathology}}}
  \textbf{\bibinfo{volume}{13}}, \bibinfo{pages}{12} (\bibinfo{year}{2018}).

\bibitem{cruz2011lung}
\bibinfo{author}{Cruz, C. S.~D.}, \bibinfo{author}{Tanoue, L.~T.} \&
  \bibinfo{author}{Matthay, R.~A.}
\newblock \bibinfo{journal}{\bibinfo{title}{Lung cancer: epidemiology,
  etiology, and prevention}}.
\newblock {\emph{\JournalTitle{Clinics in chest medicine}}}
  \textbf{\bibinfo{volume}{32}}, \bibinfo{pages}{605--644}
  (\bibinfo{year}{2011}).

\bibitem{ratcliffe2017agreement}
\bibinfo{author}{Ratcliffe, M.~J.} \emph{et~al.}
\newblock \bibinfo{journal}{\bibinfo{title}{Agreement between programmed cell
  death ligand-1 diagnostic assays across multiple protein expression cutoffs
  in non--small cell lung cancer}}.
\newblock {\emph{\JournalTitle{Clinical Cancer Research}}}
  (\bibinfo{year}{2017}).

\bibitem{kim2017pd}
\bibinfo{author}{Kim, H.}, \bibinfo{author}{Kwon, H.~J.},
  \bibinfo{author}{Park, S.~Y.}, \bibinfo{author}{Park, E.} \&
  \bibinfo{author}{Chung, J.-H.}
\newblock \bibinfo{journal}{\bibinfo{title}{Pd-l1 immunohistochemical assays
  for assessment of therapeutic strategies involving immune checkpoint
  inhibitors in non-small cell lung cancer: a comparative study}}.
\newblock {\emph{\JournalTitle{Oncotarget}}} \textbf{\bibinfo{volume}{8}},
  \bibinfo{pages}{98524} (\bibinfo{year}{2017}).

\bibitem{Ventana2017}
\bibinfo{author}{Ventana Medical System~Inc., R.~D.}
\newblock \bibinfo{journal}{\bibinfo{title}{Ventana pd-l1 (sp263) assay
  staining of non-small cell lung cancer - interpretation guide}}.
\newblock {\emph{\JournalTitle{www.ventana.com}}}  (\bibinfo{year}{2016}).

\bibitem{tsao2018pd}
\bibinfo{author}{Tsao, M.~S.} \emph{et~al.}
\newblock \bibinfo{journal}{\bibinfo{title}{Pd-l1 immunohistochemistry
  comparability study in real-life clinical samples: results of blueprint phase
  2 project}}.
\newblock {\emph{\JournalTitle{Journal of Thoracic Oncology}}}
  (\bibinfo{year}{2018}).

\bibitem{krizhevsky2012imagenet}
\bibinfo{author}{Krizhevsky, A.}, \bibinfo{author}{Sutskever, I.} \&
  \bibinfo{author}{Hinton, G.~E.}
\newblock \bibinfo{title}{Imagenet classification with deep convolutional
  neural networks}.
\newblock In \emph{\bibinfo{booktitle}{Advances in neural information
  processing systems}}, \bibinfo{pages}{1097--1105} (\bibinfo{year}{2012}).

\bibitem{simonyan2014very}
\bibinfo{author}{Simonyan, K.} \& \bibinfo{author}{Zisserman, A.}
\newblock \bibinfo{journal}{\bibinfo{title}{Very deep convolutional networks
  for large-scale image recognition}}.
\newblock {\emph{\JournalTitle{arXiv preprint arXiv:1409.1556}}}
  (\bibinfo{year}{2014}).

\bibitem{he2015delving}
\bibinfo{author}{He, K.}, \bibinfo{author}{Zhang, X.}, \bibinfo{author}{Ren,
  S.} \& \bibinfo{author}{Sun, J.}
\newblock \bibinfo{title}{Delving deep into rectifiers: Surpassing human-level
  performance on imagenet classification}.
\newblock In \emph{\bibinfo{booktitle}{Proceedings of the IEEE international
  conference on computer vision}}, \bibinfo{pages}{1026--1034}
  (\bibinfo{year}{2015}).

\bibitem{szegedy2015going}
\bibinfo{author}{Szegedy, C.} \emph{et~al.}
\newblock \bibinfo{title}{Going deeper with convolutions}
  (\bibinfo{organization}{Cvpr}, \bibinfo{year}{2015}).

\bibitem{he2016deep}
\bibinfo{author}{He, K.}, \bibinfo{author}{Zhang, X.}, \bibinfo{author}{Ren,
  S.} \& \bibinfo{author}{Sun, J.}
\newblock \bibinfo{title}{Deep residual learning for image recognition}.
\newblock In \emph{\bibinfo{booktitle}{Proceedings of the IEEE conference on
  computer vision and pattern recognition}}, \bibinfo{pages}{770--778}
  (\bibinfo{year}{2016}).

\bibitem{szegedy2016rethinking}
\bibinfo{author}{Szegedy, C.}, \bibinfo{author}{Vanhoucke, V.},
  \bibinfo{author}{Ioffe, S.}, \bibinfo{author}{Shlens, J.} \&
  \bibinfo{author}{Wojna, Z.}
\newblock \bibinfo{title}{Rethinking the inception architecture for computer
  vision}.
\newblock In \emph{\bibinfo{booktitle}{Proceedings of the IEEE Conference on
  Computer Vision and Pattern Recognition}}, \bibinfo{pages}{2818--2826}
  (\bibinfo{year}{2016}).

\bibitem{Litjens2016}
\bibinfo{author}{Litjens, G.} \emph{et~al.}
\newblock \bibinfo{journal}{\bibinfo{title}{Deep learning as a tool for
  increased accuracy and efficiency of histopathological diagnosis}}.
\newblock {\emph{\JournalTitle{Scientific reports}}}
  \textbf{\bibinfo{volume}{6}}, \bibinfo{pages}{26286} (\bibinfo{year}{2016}).

\bibitem{CruzRoa217}
\bibinfo{author}{Cruz-Roa, A.} \emph{et~al.}
\newblock \bibinfo{journal}{\bibinfo{title}{Accurate and reproducible invasive
  breast cancer detection in whole-slide images: A deep learning approach for
  quantifying tumor extent}}.
\newblock {\emph{\JournalTitle{Scientific reports}}}
  \textbf{\bibinfo{volume}{7}}, \bibinfo{pages}{46450} (\bibinfo{year}{2017}).

\bibitem{Han2017}
\bibinfo{author}{Han, Z.} \emph{et~al.}
\newblock \bibinfo{journal}{\bibinfo{title}{Breast cancer multi-classification
  from histopathological images with structured deep learning model}}.
\newblock {\emph{\JournalTitle{Scientific reports}}}
  \textbf{\bibinfo{volume}{7}}, \bibinfo{pages}{4172} (\bibinfo{year}{2017}).

\bibitem{Komura2018}
\bibinfo{author}{Komura, D.} \& \bibinfo{author}{Ishikawa, S.}
\newblock \bibinfo{journal}{\bibinfo{title}{Machine learning methods for
  histopathological image analysis}}.
\newblock {\emph{\JournalTitle{arXiv preprint arXiv:1709.00786}}}
  (\bibinfo{year}{2017}).

\bibitem{lecun1998gradient}
\bibinfo{author}{LeCun, Y.}, \bibinfo{author}{Bottou, L.},
  \bibinfo{author}{Bengio, Y.} \& \bibinfo{author}{Haffner, P.}
\newblock \bibinfo{journal}{\bibinfo{title}{Gradient-based learning applied to
  document recognition}}.
\newblock {\emph{\JournalTitle{Proceedings of the IEEE}}}
  \textbf{\bibinfo{volume}{86}}, \bibinfo{pages}{2278--2324}
  (\bibinfo{year}{1998}).

\bibitem{lecun2010convolutional}
\bibinfo{author}{LeCun, Y.}, \bibinfo{author}{Kavukcuoglu, K.} \&
  \bibinfo{author}{Farabet, C.}
\newblock \bibinfo{title}{Convolutional networks and applications in vision}.
\newblock In \emph{\bibinfo{booktitle}{Circuits and Systems (ISCAS),
  Proceedings of 2010 IEEE International Symposium on}},
  \bibinfo{pages}{253--256} (\bibinfo{organization}{IEEE},
  \bibinfo{year}{2010}).

\bibitem{Bychkov2018}
\bibinfo{author}{Bychkov, D.} \emph{et~al.}
\newblock \bibinfo{journal}{\bibinfo{title}{Deep learning based tissue analysis
  predicts outcome in colorectal cancer}}.
\newblock {\emph{\JournalTitle{Scientific reports}}}
  \textbf{\bibinfo{volume}{8}}, \bibinfo{pages}{3395} (\bibinfo{year}{2018}).

\bibitem{campanella2018terabyte}
\bibinfo{author}{Campanella, G.}, \bibinfo{author}{Silva, V. W.~K.} \&
  \bibinfo{author}{Fuchs, T.~J.}
\newblock \bibinfo{journal}{\bibinfo{title}{Terabyte-scale deep multiple
  instance learning for classification and localization in pathology}}.
\newblock {\emph{\JournalTitle{arXiv preprint arXiv:1805.06983}}}
  (\bibinfo{year}{2018}).

\bibitem{Vandenberghe2017}
\bibinfo{author}{Vandenberghe, M.~E.} \emph{et~al.}
\newblock \bibinfo{journal}{\bibinfo{title}{Relevance of deep learning to
  facilitate the diagnosis of her2 status in breast cancer}}.
\newblock {\emph{\JournalTitle{Scientific reports}}}
  \textbf{\bibinfo{volume}{7}}, \bibinfo{pages}{45938} (\bibinfo{year}{2017}).

\bibitem{zhang2016augmenting}
\bibinfo{author}{Zhang, Y.}, \bibinfo{author}{Lee, K.} \& \bibinfo{author}{Lee,
  H.}
\newblock \bibinfo{title}{Augmenting supervised neural networks with
  unsupervised objectives for large-scale image classification}.
\newblock In \emph{\bibinfo{booktitle}{International Conference on Machine
  Learning}}, \bibinfo{pages}{612--621} (\bibinfo{year}{2016}).

\bibitem{Peikari2018}
\bibinfo{author}{Peikari, M.}, \bibinfo{author}{Salama, S.},
  \bibinfo{author}{Nofech-Mozes, S.} \& \bibinfo{author}{Martel, A.~L.}
\newblock \bibinfo{journal}{\bibinfo{title}{A cluster-then-label
  semi-supervised learning approach for pathology image classification}}.
\newblock {\emph{\JournalTitle{Scientific reports}}}
  \textbf{\bibinfo{volume}{8}}, \bibinfo{pages}{7193} (\bibinfo{year}{2018}).

\bibitem{kieffer2017}
\bibinfo{author}{Kieffer, B.}, \bibinfo{author}{Babaie, M.},
  \bibinfo{author}{Kalra, S.} \& \bibinfo{author}{Tizhoosh, H.}
\newblock \bibinfo{journal}{\bibinfo{title}{Convolutional neural networks for
  histopathology image classification: Training vs. using pre-trained
  networks}}.
\newblock {\emph{\JournalTitle{arXiv preprint arXiv:1710.05726}}}
  (\bibinfo{year}{2017}).

\bibitem{mormont2018}
\bibinfo{author}{Mormont, R.}, \bibinfo{author}{Geurts, P.} \&
  \bibinfo{author}{Mar{\'e}e, R.}
\newblock \bibinfo{title}{Comparison of deep transfer learning strategies for
  digital pathology}.
\newblock In \emph{\bibinfo{booktitle}{2018 IEEE/CVF Conference on Computer
  Vision and Pattern Recognition Workshops (CVPRW)}}
  (\bibinfo{organization}{IEEE}, \bibinfo{year}{2018}).

\bibitem{Sparks2016}
\bibinfo{author}{Sparks, R.} \& \bibinfo{author}{Madabhushi, A.}
\newblock \bibinfo{journal}{\bibinfo{title}{Out-of-sample extrapolation
  utilizing semi-supervised manifold learning (ose-ssl): Content based image
  retrieval for histopathology images}}.
\newblock {\emph{\JournalTitle{Scientific reports}}}
  \textbf{\bibinfo{volume}{6}}, \bibinfo{pages}{27306} (\bibinfo{year}{2016}).

\bibitem{miyato2018spectral}
\bibinfo{author}{Miyato, T.}, \bibinfo{author}{Kataoka, T.},
  \bibinfo{author}{Koyama, M.} \& \bibinfo{author}{Yoshida, Y.}
\newblock \bibinfo{journal}{\bibinfo{title}{Spectral normalization for
  generative adversarial networks}}.
\newblock {\emph{\JournalTitle{arXiv preprint arXiv:1802.05957}}}
  (\bibinfo{year}{2018}).

\bibitem{odena2016conditional}
\bibinfo{author}{Odena, A.}, \bibinfo{author}{Olah, C.} \&
  \bibinfo{author}{Shlens, J.}
\newblock \bibinfo{journal}{\bibinfo{title}{Conditional image synthesis with
  auxiliary classifier gans}}.
\newblock {\emph{\JournalTitle{arXiv preprint arXiv:1610.09585}}}
  (\bibinfo{year}{2016}).

\bibitem{goodfellow2014}
\bibinfo{author}{Goodfellow, I.} \emph{et~al.}
\newblock \bibinfo{title}{Generative adversarial nets}.
\newblock In \emph{\bibinfo{booktitle}{Advances in neural information
  processing systems}}, \bibinfo{pages}{2672--2680} (\bibinfo{year}{2014}).

\bibitem{bug2017analyzing}
\bibinfo{author}{Bug, D.}, \bibinfo{author}{Grote, A.},
  \bibinfo{author}{Sch{\"u}ler, J.}, \bibinfo{author}{Feuerhake, F.} \&
  \bibinfo{author}{Merhof, D.}
\newblock \bibinfo{title}{Analyzing immunohistochemically stained whole-slide
  images of ovarian carcinoma}.
\newblock In \emph{\bibinfo{booktitle}{Bildverarbeitung f{\"u}r die Medizin
  2017}}, \bibinfo{pages}{173--178} (\bibinfo{publisher}{Springer},
  \bibinfo{year}{2017}).

\bibitem{Otsu1979}
\bibinfo{author}{Otsu, N.}
\newblock \bibinfo{journal}{\bibinfo{title}{A threshold selection method from
  gray-level histograms}}.
\newblock {\emph{\JournalTitle{IEEE transactions on systems, man, and
  cybernetics}}} \textbf{\bibinfo{volume}{9}}, \bibinfo{pages}{62--66}
  (\bibinfo{year}{1979}).

\bibitem{russell2016artificial}
\bibinfo{author}{Russell, S.~J.} \& \bibinfo{author}{Norvig, P.}
\newblock \emph{\bibinfo{title}{Artificial intelligence: a modern approach}}
  (\bibinfo{publisher}{Malaysia; Pearson Education Limited,},
  \bibinfo{year}{2016}).

\bibitem{cho2017neural}
\bibinfo{author}{Cho, H.}, \bibinfo{author}{Lim, S.}, \bibinfo{author}{Choi,
  G.} \& \bibinfo{author}{Min, H.}
\newblock \bibinfo{journal}{\bibinfo{title}{Neural stain-style transfer
  learning using gan for histopathological images}}.
\newblock {\emph{\JournalTitle{arXiv preprint arXiv:1710.08543}}}
  (\bibinfo{year}{2017}).

\bibitem{Shaban2018}
\bibinfo{author}{Shaban, M.~T.}, \bibinfo{author}{Baur, C.},
  \bibinfo{author}{Navab, N.} \& \bibinfo{author}{Albarqouni, S.}
\newblock \bibinfo{journal}{\bibinfo{title}{Staingan: Stain style transfer for
  digital histological images}}.
\newblock {\emph{\JournalTitle{arXiv preprint arXiv:1804.01601}}}
  (\bibinfo{year}{2018}).

\bibitem{chen2016infogan}
\bibinfo{author}{Chen, X.} \emph{et~al.}
\newblock \bibinfo{title}{Infogan: Interpretable representation learning by
  information maximizing generative adversarial nets}.
\newblock In \emph{\bibinfo{booktitle}{Advances in Neural Information
  Processing Systems}}, \bibinfo{pages}{2172--2180} (\bibinfo{year}{2016}).

\bibitem{SzegedyVISW15}
\bibinfo{author}{Szegedy, C.}, \bibinfo{author}{Vanhoucke, V.},
  \bibinfo{author}{Ioffe, S.}, \bibinfo{author}{Shlens, J.} \&
  \bibinfo{author}{Wojna, Z.}
\newblock \bibinfo{journal}{\bibinfo{title}{Rethinking the inception
  architecture for computer vision}}.
\newblock {\emph{\JournalTitle{CoRR}}}
  \textbf{\bibinfo{volume}{abs/1512.00567}} (\bibinfo{year}{2015}).
\newblock \eprint{1512.00567}.

\bibitem{biglenet}
\bibinfo{author}{LeCun, Y.}, \bibinfo{author}{Bottou, L.},
  \bibinfo{author}{Bengio, Y.} \& \bibinfo{author}{Haffner, P.}
\newblock \bibinfo{journal}{\bibinfo{title}{Gradient-based learning applied to
  document recognition}}.
\newblock {\emph{\JournalTitle{Proceedings of the IEEE}}}
  \textbf{\bibinfo{volume}{86}}, \bibinfo{pages}{2278--2324}
  (\bibinfo{year}{1998}).

\bibitem{ioffe2015batch}
\bibinfo{author}{Ioffe, S.} \& \bibinfo{author}{Szegedy, C.}
\newblock \bibinfo{journal}{\bibinfo{title}{Batch normalization: Accelerating
  deep network training by reducing internal covariate shift}}.
\newblock {\emph{\JournalTitle{arXiv preprint arXiv:1502.03167}}}
  (\bibinfo{year}{2015}).

\bibitem{hinton2006reducing}
\bibinfo{author}{Hinton, G.~E.} \& \bibinfo{author}{Salakhutdinov, R.~R.}
\newblock \bibinfo{journal}{\bibinfo{title}{Reducing the dimensionality of data
  with neural networks}}.
\newblock {\emph{\JournalTitle{science}}} \textbf{\bibinfo{volume}{313}},
  \bibinfo{pages}{504--507} (\bibinfo{year}{2006}).

\bibitem{rebelatto2016development}
\bibinfo{author}{Rebelatto, M.~C.} \emph{et~al.}
\newblock \bibinfo{journal}{\bibinfo{title}{Development of a programmed cell
  death ligand-1 immunohistochemical assay validated for analysis of non-small
  cell lung cancer and head and neck squamous cell carcinoma}}.
\newblock {\emph{\JournalTitle{Diagnostic pathology}}}
  \textbf{\bibinfo{volume}{11}}, \bibinfo{pages}{95} (\bibinfo{year}{2016}).

\bibitem{kingma2014adam}
\bibinfo{author}{Kingma, D.~P.} \& \bibinfo{author}{Ba, J.}
\newblock \bibinfo{journal}{\bibinfo{title}{Adam: A method for stochastic
  optimization}}.
\newblock {\emph{\JournalTitle{arXiv preprint arXiv:1412.6980}}}
  (\bibinfo{year}{2014}).

\bibitem{chollet2015keras}
\bibinfo{author}{Chollet, F.} \emph{et~al.}
\newblock \bibinfo{title}{Keras}.
\newblock \bibinfo{howpublished}{\url{https://keras.io}}
  (\bibinfo{year}{2015}).

\bibitem{tensorflow2015-whitepaper}
\bibinfo{author}{Abadi, M.} \emph{et~al.}
\newblock \bibinfo{title}{{TensorFlow}: Large-scale machine learning on
  heterogeneous systems} (\bibinfo{year}{2015}).
\newblock \bibinfo{note}{Software available from tensorflow.org}.

\end{thebibliography}

\end{document}